\documentclass{article}
\usepackage{arxiv}

\usepackage[utf8]{inputenc} % allow utf-8 input
\usepackage[T1]{fontenc}    % use 8-bit T1 fonts
\usepackage{hyperref}       % hyperlinks
\usepackage{url}            % simple URL typesetting
\usepackage{datetime}       % show dates in the title block
\usepackage{booktabs}       % professional-quality tables
\usepackage{amsfonts}       % blackboard math symbols
\usepackage{nicefrac}       % compact symbols for 1/2, etc.
\usepackage{microtype}      % microtypography
\usepackage{graphicx}
\usepackage{natbib}
\usepackage{doi}
\usepackage{xcolor}

\usepackage{amsmath}
\hypersetup{colorlinks = true,
linkcolor = purple,
urlcolor  = blue,
citecolor = cyan,
anchorcolor = black}

\title{Biomechanical Reconstruction with Confidence Intervals from Multiview Markerless Motion Capture}

\makeatletter
\let\@fnsymbol\@arabic
\makeatother

\author{\href{https://orcid.org/0000-0001-5714-1400}{\includegraphics[scale=0.06]{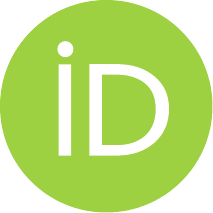}}\hspace{1mm}R. James Cotton\footnotemark[1]\\
Shirley Ryan AbilityLab\\Northwestern University\\\AND
Fabian Sinz\\
University of Göttingen\\}

\renewcommand{\headeright}{}
\renewcommand{\undertitle}{}
\renewcommand{\shorttitle}{SSL Biomechanical Uncertainty}

\hypersetup{
pdftitle={\@title},
pdfsubject={},
pdfauthor={\@author},
pdfkeywords={c,o,m,p,u,t,e,r, ,v,i,s,n,,,a,y,l,b,h,d,g},
addtopdfcreator={Written in Curvenote}
}

\date{}  

\begin{document}
\maketitle
\footnotetext[1]{Correspondence to: rcotton@sralab.org}

\begin{abstract}
Advances in multiview markerless motion capture (MMMC) promise high-quality movement analysis for clinical practice and research. While prior validation studies show MMMC performs well on average, they do not provide what is needed in clinical practice or for large-scale utilization of MMMC -- confidence intervals over specific kinematic estimates from a specific individual analyzed using a possibly unique camera configuration. We extend our previous work using an implicit representation of trajectories optimized end-to-end through a differentiable biomechanical model to learn the posterior probability distribution over pose given all the detected keypoints. This posterior probability is learned through a variational approximation and estimates confidence intervals for individual joints at each moment in a trial, showing confidence intervals generally within 10-15 mm of spatial error for virtual marker locations, consistent with our prior validation studies. Confidence intervals over joint angles are typically only a few degrees and widen for more distal joints. The posterior also models the correlation structure over joint angles, such as correlations between hip and pelvis angles. The confidence intervals estimated through this method allow us to identify times and trials where kinematic uncertainty is high.
\end{abstract}

\section{Introduction}

Accurate and accessible human movement analysis holds great promise for research and medicine. However, two capabilities are essential to fulfilling this promise are 1) measuring clinically meaningful features, such as biomechanically grounded joint angles, and 2) providing valid confidence intervals. The latter is particularly important, as it is equally important to know whether a measurement can be trusted as it is to know the measured value.

Recently, we showed that differentiable biomechanical models can improve biomechanical reconstruction from multiview markerless motion capture (MMC) \citep{cotton_differentiable_2024}. This was made possible through the recent support for GPU acceleration of biomechanical models in the MuJoCo physics simulator \citep{todorov_mujoco_2012}, as well as work from the MyoSuite project \citep{caggiano_myosuite_2022} improving these models. We fit biomechanical models using an implicit function that mapped from time to joint configurations, then passed the joint configurations through the forward kinematic model of the body and reprojected this body model through the camera model. The reprojection error between the predicted marker locations and the detected keypoints is minimized by optimizing the implicit trajectory model. This end-to-end approach produced lower reprojection errors than a traditional multistage biomechanical reconstruction pipeline that first reconstructed virtual marker trajectories and then performed inverse kinematics on these trajectories.

However, neither a traditional nor end-to-end approach provides measures of their confidence bounds over joint angles. With markerless motion capture, this can vary due to keypoint noise, occlusions, and camera geometry. Common approaches to validating markerless motion capture characterize the average performance of a particular camera configuration for specific measurements on a specific human population (e.g., a clinical diagnosis on able-bodied controls). However, this does not provide the information truly needed -- how accurate is a particular measurement with a novel camera configuration on the next current participant?

Multiview geometry provides cues to inform the uncertainty over a specific trajectory. If a marker location is observed by more than 2 calibrated cameras, the Direct Linear Transform (DLT) equations for describing this mapping are overdetermined \citep{hartley_multiple_2003}. Any inconsistencies, i.e., reprojection errors, after triangulating the 3D locations may arise from the keypoint detection errors, which can be signalled by the confidence score from the keypoint detector can also signal. These errors should be reflected in a larger spatial confidence interval around the inferred 3D location, ultimately producing a wider uncertainty over the joint angle inferred through inverse kinematics. Biomechanics also provide additional constraints, such as joint angle limits, constant bone lengths, and the relative configuration of different virtual markers on a given body segment remaining constant. Any inconsistencies in these constraints should result in a trajectory with greater uncertainty. However, these information sources have not been leveraged to infer the uncertainty over biomechanical trajectories.

\begin{figure}[!htbp]
\centering
\includegraphics[width=1\linewidth]{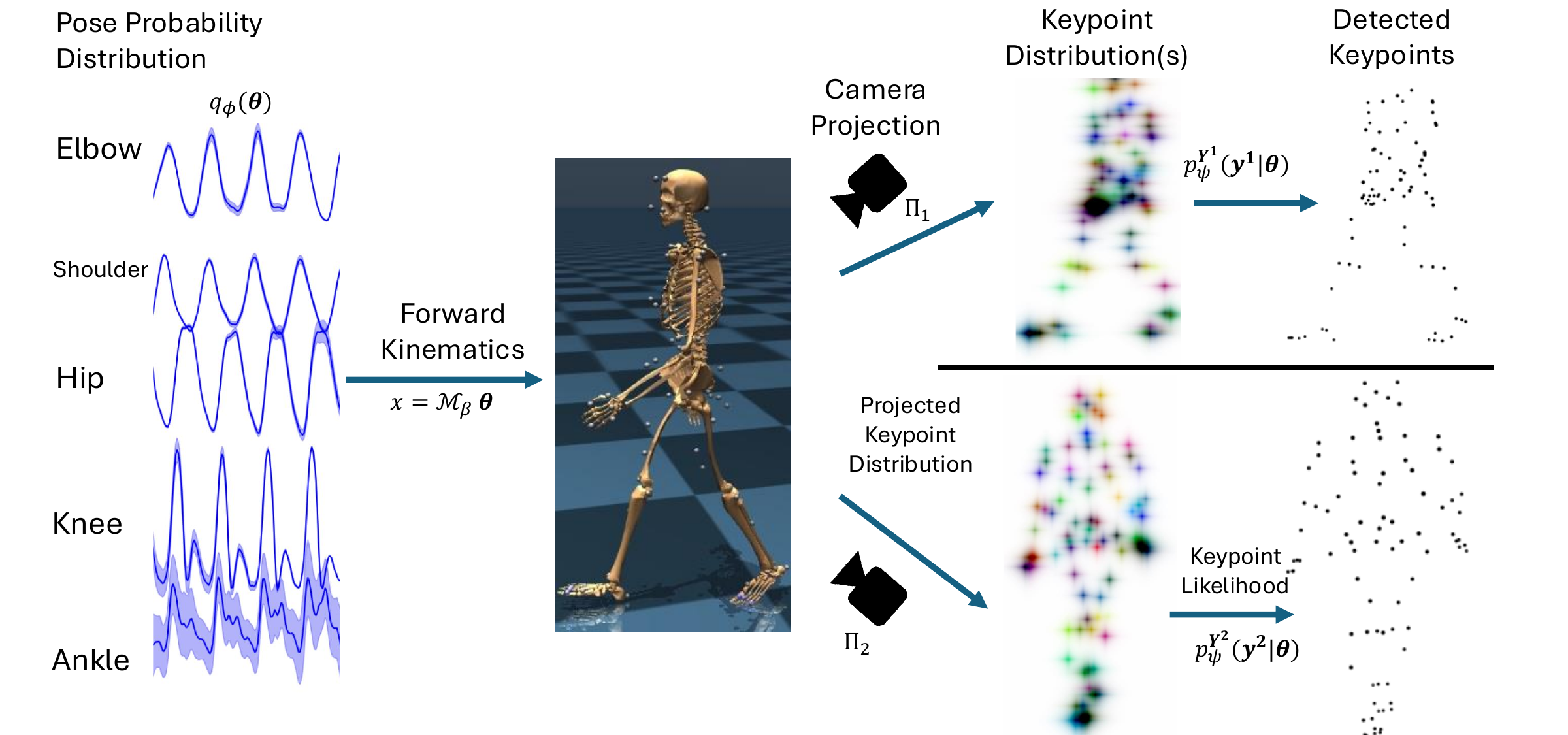}
\caption[]{Overview of our approach. Samples from the variational posterior distribution, $q_\phi(\boldsymbol \theta)$ is transformed by the biomechanical forward kinematic model, $\mathcal M_\beta$ and then projected through the camera model $\Pi_c$ to produce a predicted probability distribution over the keypoints, $p_\phi^{\mathbf Y^c}(\mathbf y^c|\boldsymbol \theta)$. The detected keypoints are evaluated by this likelihood function to optimize the ELBO.}
\label{fig:biomechanical_uncertainty_overview}
\end{figure}

In this work, we show how the implicit representation over a trajectory can be extended to a probabilistic representation. We then show how this can fit synchronized multiview videos using variational inference. This is made more challenging as keypoint detectors are not designed to be probabilistic and, thus, do not have a well-characterized likelihood function over their detections. However, this can be jointly learned during the inference process. Finally, we show that these confidence intervals are well-calibrated. This algorithm results in per-trial confidence intervals over the individual joints at each moment of the trial, allowing data to be discarded when occlusions or camera geometry make the data segments unreliable. Figure~\ref{fig:biomechanical_uncertainty_overview} shows an overview of our approach.

The current work addresses these issues with the following key contributions:

\begin{itemize}
\item We fit an implicit representation for a probability distribution over kinematic joint angle trajectories to markerless motion capture data that includes a low-rank correlation structure.
\item We optimize this end-to-end using a fast, differentiable biomechanical model to match the detected keypoints.
\item We show that this produces well-calibrated estimates of uncertainty while providing tight confidence intervals over joint angles.
\end{itemize}

\section{Method}

\subsection{Variational Inference of Kinematic Trajectory Uncertainty}

\subsubsection{Preliminaries}

The multiview markerless motion capture system produces a sequence of images for each camera at each point in time. These are processed by a keypoint detector that outputs a set of keypoints and confidence scores for each image: $(\mathbf Y_t, \mathbf S_t) = \left\{ (\mathbf y_t^c, \mathbf s_t^c) \, \forall \, c \right\}$. Our goal is to compute the posterior over the pose parameters for each point in time $p(\boldsymbol \theta_t | \mathbf Y_t, \mathbf S_t)$, where $\boldsymbol \theta \in \mathbb R^{40}$ are the kinematic joint angles at time $t$.

The pose parameters predict the detected keypoint locations through two non-linear steps, described in greater depth in \citet{cotton_differentiable_2024}. First, the marker locations can be predicted in 3D space from the pose parameter using the forward kinematic model of the body, $\mathcal M_\beta: \boldsymbol \theta \rightarrow \mathbf x$, where $\mathbf x \in \mathbb R^{87 \times 3}$ are a set of dense virtual markers that cover the body. The personalized model scale is $\beta \in \mathbb R^{8+J*3}$, which includes 8 body scaling parameters and $\mathbf x \in \mathbb R^{J \times 3}$ are the $J=87$ predicted marker locations.

Next, these marker locations are projected into the image plane using the calibrated camera model, which includes full perspective and distortion parameters, $\Pi_c: \mathbf x \rightarrow \hat{\mathbf y}^c$. Throughout this work, we will use the notation $\hat{\mathbf y}^c$ to distinguish hypothesized predictions of marker locations from the actual keypoints detected from the images, $\mathbf y^c$.

In principle, we could compute the posterior of the pose trajectory given the observations using Bayes' rule: $p(\boldsymbol \theta_t | \mathbf Y_t, \mathbf S_t) = \frac{p(\mathbf Y_t, \mathbf S_t | \boldsymbol \theta_t) p(\boldsymbol \theta_t)}{p(\mathbf Y_t, \mathbf S_t)}$. However, this is challenging because we do not have the normalization factor $p(\mathbf Y_t, \mathbf S_t)$, nor is it tractable to obtain it by marginalizing over the high-dimension pose parameters. Another challenge is that the likelihood function $p(\mathbf Y_t, \mathbf S_t | \boldsymbol \theta_t)$ is not directly available. Even though we can hypothesize where the ground truth keypoints may be on the image plane, $\hat{\mathbf y_t^c} = \Pi_c \mathcal M_\beta \boldsymbol \theta_t$, we do not have a likelihood function over detected locations and confidence scores, $p(\mathbf y_t^c, \mathbf s_t^c | \hat{\mathbf y_t^c})$. In particular, the keypoint confidence scores of most algorithms are not calibrated to parameterize such a likelihood function.

\subsubsection{Variational Approximation}

We follow the common variational inference approach to estimate the posterior despite the intractable normalization. Specifically, we approximate the posterior with a Gaussian distribution, $q(\boldsymbol \theta_t)$, and then minimize the Kullback-Leibler (KL) divergence between this approximation and the true posterior: $D_{\text{KL}}[q(\boldsymbol \theta_t) || p(\boldsymbol \theta_t | \mathbf Y_t, \mathbf S_t)]$. While the true posterior is not available, minimizing a lower bound on the KL divergence is possible. We have dropped the explicit notation of time for notational clarity, but note this is optimized for every timestep.
\begin{align*}
\min D_{KL}[q(\boldsymbol \theta) || p(\boldsymbol \theta|\mathbf Y, \mathbf S)]
 & = \int_q q(\boldsymbol \theta) \log \frac{q(\boldsymbol \theta)}{p(\boldsymbol \theta | \mathbf Y, \mathbf S)} \\
  &= \mathbb E_{q(\boldsymbol \theta)} \left[ \log \frac{q(\boldsymbol \theta)}{p(\boldsymbol \theta | \mathbf Y, \mathbf S)} \right] \\
 &= \underbrace{\mathbb E_{q(\boldsymbol \theta)} \left[ \log q(\boldsymbol \theta) - \log p(\boldsymbol \theta , \mathbf Y, \mathbf S) \right]}_{\text{-ELBO}} + \log p(\mathbf Y, \mathbf S)
 \end{align*}

Because $p(\mathbf Y, \mathbf S)$ is a constant, we can maximize the Evidence Lower Bound (ELBO) to minimize the KL divergence:
\begin{align}
\text{ELBO}(q) : &= \mathbb E_{q(\boldsymbol \theta)} \left[-\log q(\boldsymbol \theta) + \log p(\mathbf Y, \mathbf S | \boldsymbol \theta) + \log p(\boldsymbol \theta)  \right] \nonumber \\
&= H(q) + \mathbb E_{q \sim q(\boldsymbol \theta)} \left[ \log p(\mathbf Y, \mathbf S | \boldsymbol \theta )  + \log p(\boldsymbol\theta) \right] \label{elbo}
\end{align}

The prior over poses, $p(\boldsymbol \theta)$, could be learned from data to reflect common body positions, which would bias solutions towards these postures. However, because we are primarily targeting clinical populations who may have movement differences from the able-bodied population, we use an unnormalized uniform distribution over the joint limits of our model and a very low value for values outside this range. Below, we elaborate on the likelihood function over detected keypoints, $p(\mathbf Y, \mathbf S | \boldsymbol \theta )$.

Thus, at a high level, it is clear that maximizing the ELBO will be inclined to increase the entropy of the estimated posterior distribution over the pose (i.e., will produce greater uncertainty over joint angles), which may be counterbalanced by evidence from the detected keypoints to reduce it.

\subsubsection{Implicit Representation of Probability Distribution over Pose}

Variational inference requires a representation of the probability distribution over the pose trajectory, $q(\boldsymbol \theta_t)$, that can be optimized. Following \citet{cotton_differentiable_2024}, we use an implicit function that maps from time to the trajectory and extend it to output both the mean and covariance of the trajectory: $f_\phi: t \rightarrow (\boldsymbol \mu, \mathbf u)$. $f_\phi$ is implemented using an MLP that outputs a vector for $\boldsymbol \mu_{\boldsymbol \theta} \in \mathbb R^{40}$ and $\mathbf u \in \mathbb R^{40 \times R}$. $\mathbf u$ parameterize a low-rank representation of $\Sigma_{\boldsymbol \theta}$, with rank $R$.
\begin{align*}
q_\phi(\boldsymbol \theta_t)=\mathcal N \left( \boldsymbol \theta_t ; \, \boldsymbol \mu_\phi(t), \Sigma_\phi(t) \right)
\end{align*}

The first 6 elements of the pose are the position and orientation of the pelvis. We sometimes use a version that uses a quaternion for orientation, but we did not use it in this writeup as the Euler parameterization produces more interpretable correlation structures. The remaining 34 elements consist of joints with biomechanical limits. When computing $\mu_\phi$, we passed those 34 elements through a $\mathrm{tanh}$ nonlinearity and rescaled them to match the limits specified by the model. Because this only constrains the mean location of the probabilistic representation, some samples may still exceed the joint limits. However, these are further penalized by the pose prior, $p(\boldsymbol \theta)$.

\subsubsection{Likelihood Function}

As noted above, keypoint detectors do not come with calibrated likelihood functions, which we need for computing $p(\mathbf Y, \mathbf S | \boldsymbol \theta )$ in Eq (\ref{elbo}). Our forward kinematic model and camera calibration allow us to predict where the keypoint would be detected for a given pose: $\hat{\mathbf y}^c = \Pi_c \mathcal M_\beta \boldsymbol \theta$. Additionally, each keypoint comes with an associated scalar confidence score, $s_j^c$, which should roughly predict the keypoint accuracy. For the keypoint detector we are using, the score measures predicted noise and increases for less accurate keypoints. We treat the measured confidence score for each keypoint as independent of the predicted location, and also that the likelihood of the detected location is conditionally independent for other keypoints and only depends on the Euclidean spatial error between the predicted location and detected location, $\epsilon_j^c = ||\mathbf y_{j}^c - \hat{\mathbf y}_{j}^c ||$, as well as the confidence score.
\begin{align*}
\log p(\mathbf Y, \mathbf S | \boldsymbol \theta) &= \sum_{c,j} \log p(\mathbf y_{j}^c, s_{j}^c | \hat{\mathbf y}_{j}^c ) \\ 
&= \sum_{c,j} \log p_\psi(\mathbf y_{j}^c | \hat{\mathbf y}_{j}^c , s_{j}^c ) p(s_{j}^c) \\
&= \sum_{c,j} \log p_\psi(\epsilon_j^c, | s_{j}^c) + \log p(s_{j}^c) 
\end{align*}

We parameterize the width of the error distribution as a second-order polynomial of the confidence score, $\sigma_\psi(s) = (\mathrm{softplus}(\psi_0) + \mathrm{softplus}(\psi_1) s + \mathrm{softplus}(\psi_2) s^2)$, where the $\mathrm{softplus}$ ensures the width remains positive and only can increase with higher scores that indicate greater noise. Below, we compare several distribution types, finding the exponential is generally the best, although the half-Cauchy also works well. The shallower tails of the Half Normal distribution make learning unstable. Because Eq (\ref{elbo}) is optimized with respect to $\boldsymbol \theta$, $p(s_j^c)$ has no influence. Thus, we ultimately maximize:
\begin{equation*}
\text{ELBO}(q) : =  H(q) + \mathbb E_{q \sim q(\boldsymbol \theta)} \left[  \log p(\boldsymbol \theta)  + \sum_{c,j} \log p_\psi(||\mathbf y_{j}^c - (\Pi_c \mathcal M_\beta \boldsymbol \theta)_{j} || \, | \sigma(s_{j}^c)) \right]
\end{equation*}
The likelihood parameters $\psi$ are allowed to change with $\phi$ during the optimization.

\subsubsection{Implementation Details}

We use a high-performance, GPU-acclerated, and differentiable biomechanical model running in MuJoCo \citep{todorov_mujoco_2012} for our forward kinematics, similar to our prior work in \citep{cotton_differentiable_2024}. $p_\phi(\boldsymbol \theta_t)$ is an implicit representation implemented as an MLP,  $f_\phi$, which outputs the mean and covariance of the probability distribution for each time. The MLP had hidden layers of sizes 128, 256, 512, 1024 with $\mathrm{relu}$ nonlinearities. This was implemented in Jax and Equinox \citep{kidger_equinox_2021}. Time was scaled to not exceed $\pi$ and was positionally encoded using an encoding dimension that ensured a minimum frequency of 80 Hz. The probability function was optimized using the reparameterization trick \citep{kingma_auto-encoding_2014}, using 8 samples per gradient step. We performed optimization with the AdamW optimizer from Optax \citep{loshchilov_decoupled_2019, deepmind2020jax}, using $\beta_1=0.8$ and weight decay of $1e -5$. The learning rate used included an exponential decay from an initial value of $1e -3$ to an end value of $1e -8$. We used 30k optimization steps with the keypoint likelihood learning enabled at 2.5k using a separate Adam optimizer with a learning rate of $1e -3$. The probability distributions used implemented Tensorflow Probability \citep{dillon_tensorflow_2017}.

Bilevel optimization: Unless otherwise specified, we fit multiple trajectories from the same participant in parallel. In this case, multiple $f_\phi$ were learned to represent the implicit trajectories, but a single set of $\beta$ and $\psi$ were learned. This bilevel optimization approach of jointly estimating the skeleton scale and marker offsets while estimating in inverse kinematics stands in contrast to traditional approaches that require a static scaling trial \citep{werling_rapid_2022, cotton_differentiable_2024}. Because Jax recompiles code for different input sizes, we sampled 100 evenly spaced timesteps from each trial to compute each gradient step. As in \citet{cotton_differentiable_2024} the camera extrinsic parameters could also be refined later during the optimization process (after 10k steps), which we found slightly improved the results.  We also kept the $\psi$ parameters frozen for the first 7k steps.

\subsection{Metric: Expected Calibration Error}

A key motivation of this work are accurate confidence intervals for our pose estimates in order to indicate when and how much our data can be trusted. These confidence intervals must be well-calibrated. For example, if we estimate a 95\% confidence interval, the ground truth should fall within this interval approximately 95\% of the time. This can be quantified with the Expected Calibration Error (ECE) \citep{song_distribution_2019, pierzchlewicz_optimizing_2023}, which measures the agreement between predicted confidence intervals and the actual empirical distribution of errors. However, this requires ground truth poses, which we do not have. Instead, we measure ECE with respect to the detected keypoints, although they are not ground truth. Briefly, we transform the posterior distribution in the pose space into a posterior distribution in the image keypoints space. This transformation alone does not account for the expected variance from the keypoint likelihood function, $\sigma_\psi(s)$. We can add the variance from the learned likelihood function to the variance from the predicted posterior projected on the image plane to obtain an overall distribution for the keypoints, and then can measure the ECE of these keypoints similarly to our prior work \citep{pierzchlewicz_optimizing_2023}. To make this estimate more conservative and avoid a situation where excessive uncertainty is attributed to the likelihood function, we make two modifications. First, we clip the width of the likelihood function to a maximum value, $\sigma=\min(\sigma_\psi(s), \sigma_{\mathrm ECE})$ prior to adding it to the posterior variance and computing the ECE, which we denote as $\mathrm{ECE}_{\sigma_{\mathrm ECE}}$. Second, we only use the 5th percentile of keypoints with the highest confidence score as a pseudo-ground truth. This should have the least measurement noise and provide the best assessment of calibration.

\section{Experiments}

\subsection{Synthetic problem and method validation}

While we do not include the results here due to length restrictions, we created a synthetic toy example of this problem using multiple keypoints on a single axis measured via an affine transformation. These measurements had variable levels of noise detected that were indicated by a confidence score. These synthetic experiments ensured that combining a variational approximation of the posterior with estimating the parameters of the likelihood function both recovered the actual likelihood function. Furthermore, the estimated posterior showed well-calibrated confidence intervals.

These experiments also revealed that discrepancies in the affine transformation (equivalent to $\Pi$) used when generating the data versus those used in the variational approximation could result in the likelihood function having $\psi$ parameters that were biased positively. This allowed the model to attribute too much uncertainty to the observations and insufficient uncertainty to the posterior, producing overly confident estimates. Because our results on real data find small values for $\psi$ and we also optimize the $\Pi$ camera model, we do not believe this is a significant issue in practice. Finally, these toy examples demonstrate our approach to measuring ECE from the high-confidence measured keypoints were reliable even in the presence of this model misspecification, providing additional verification.

\subsection{Participants}

The Northwestern University Institutional Review Board approved this study. We collected data from a diverse population of individuals seen for rehabilitation at Shirley Ryan AbilityLab as either inpatients or outpatients. This included individuals who use a lower limb prosthesis, individuals with gait impairments from a neurologic condition, and able-bodied individuals. This data was collected in multiple locations using cameras installed in a laboratory, cameras set up temporarily with tripods along a hallway, and a setup installed in a linear pattern on the roof of a therapy gym. Some data was also collected at conferences, including the 2024 AOPA conference, where we collected data from able-bodied individuals and lower limb prosthesis users, and the 2024 ASB conference, where we collected data from able-bodied individuals. Thus, throughout this study, we use a wide variety of camera configurations and spatial scales. These sessions primarily focused on mobility studies such as walking, timed up and go, the four-square step test, and the functional gait assessment (excluding stairs).

\subsection{Data Collection and Preprocessing}

Data was collected and preprocessed as described in \citet{cotton_differentiable_2024}. In brief, synchronized multiview video was acquired using our custom software from 8 to 12 FLIR BlackFly S GigE cameras. Checkerboards or charuco patterns were used to calibrate the camera's intrinsic and extrinsic parameters. Easymocap was used for initial scene reconstruction and segmentation \citep{dong2021fast}, and a custom visualization tool was used to annotate the person undergoing analysis. Images were processed with the MetrABs-ACAE algorithm \citep{sarandi_learning_2023}, which is trained on the superset of keypoints used in 3D datasets, and we retained the 87 keypoints from the MOVI dataset \citep{ghorbani_movi_2021}. Our prior work found these keypoints work well for biomechanics, with the dense coverage of the whole body stabilizing estimates of the trunk and torso \citep{cotton_optimizing_2023, cotton_improved_2023}. Our tracking algorithm sometimes includes keypoints from people between the participant and the camera, such as physical therapists providing assistance. We use a robust triangulation on the keypoints to identify cameras associated with these outliers and remove them from the analysis \citep{cotton_improved_2023}.

\subsection{Method Optimization and Validation}

There are numerous hyperparameters in our method. We systematically varied several of these to understand their impact on the results. These experiments included changing the functional form of the likelihood function and the distribution rank. To further validate our method, we tested the impact of synthetically removing cameras and injecting keypoint noise.  Because it took a significant amount of computing time to run all these variations, we ran them on a set of sessions that captured some of the diversity of our applications. This includes varied camera geometries and participants (able-bodied, lower-limb prosthesis users, and stroke survivors). Hyperparameter and lesion experiments were tracked using Weights and Biases \citep{wandb}. In these cases, we used only four trials from a given session.

\subsubsection{Likelihood Function}

We compared three different forms for the likelihood function of the non-negative spatial errors, $p(\epsilon_j^c | \sigma(s_{j}^c))$: a Half Normal, an Exponential, and a Half Cauchy distribution. These have increasingly heavy tails, which can make the result less sensitive to outliers. To compare the methods, we used the ELBO as our primary metric. This is not perfect, as it is a lower bound on the true likelihood. However, measuring the true log-likelihood under the different models is intractable, which motivates our variational approximation.

We found that the Half Normal optimization diverged, likely due to the shallow tails producing extremely low likelihoods during initialization and unstable gradients. The exponential and half-Cauchy produced similar ELBOs, but the ELBO was consistently slightly higher for the exponential, so we used that for subsequent experiments.

\subsubsection{Distribution Rank}

We also tested different ranks for the probability distribution (specifically 0, 1, 2, 5, 10, 20, 30, and 40). We found the ELBO was fairly similar across these values, as were the mean reprojection errors. However, the entropy of the distribution did increase slightly for higher ranks, with a corresponding small increase in the average joint uncertainty. There was also a slight decrease in $\mathrm{ECE}_0$ with higher rank distributions. This effect plateaued around a rank of 20, which we used for the remainder of our experiments.

\begin{figure}[!htbp]
\centering
\includegraphics[width=1\linewidth]{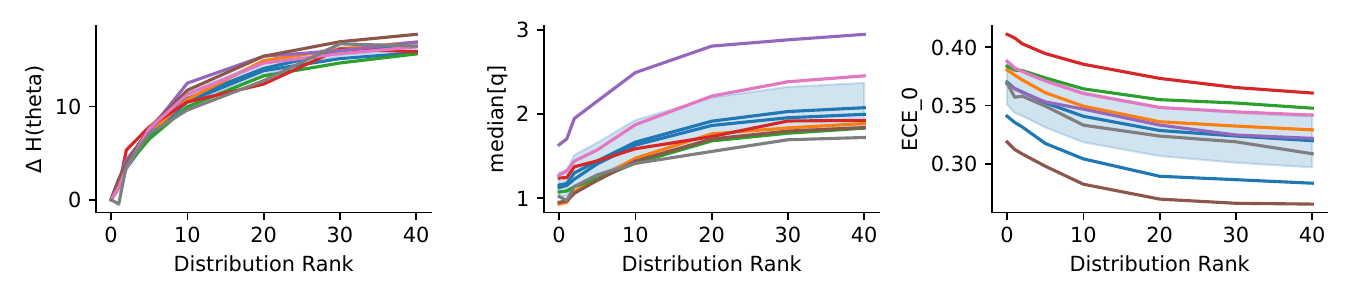}
\caption[]{Influence of distribution rank. The left panel shows the distribution entropy relative to the 0-rank distribution. The middle shows the median value for the joint angle uncertainty, taken over time and joints. The right panel shows the $\mathrm{ECE}_0$ for the different ranks.}
\label{fig:distribution_rank}
\end{figure}

\subsubsection{Injection Keypoint Noise}

To validate that our method correctly estimates the keypoint noise level and to test the influence of keypoint noise, we artificially injected additional noise on the keypoints without altering the confidence scores. This caused the entropy and median joint angle error to increase. The value of $\psi_0$ also increased and closely tracked the injected noise level with slightly lower noises due to the actual noise.

\begin{figure}[!htbp]
\centering
\includegraphics[width=1\linewidth]{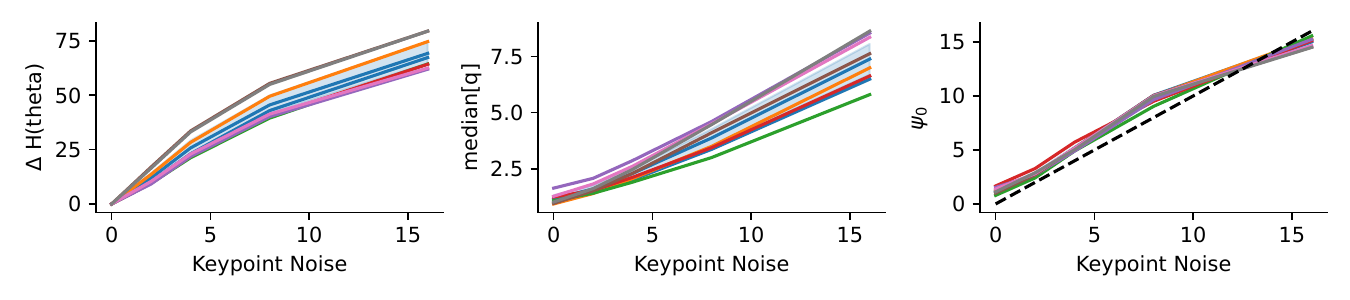}
\caption[]{Influence of artificially injected keypoint noise. The left panel shows the distribution entropy relative to the no-noise condition. The middle shows the median value for the joint angle uncertainty. The right panel shows $\psi_0$ for the different noise levels.}
\label{fig:keypoint_noise}
\end{figure}

\subsubsection{Removing Cameras}

We also reanalyzed sessions using only a subset of the cameras, down to two. To make this analysis internally consistent, we only used time points where the person was visible from both cameras when we were only using 2. Notably, our method handles the condition with only one or no cameras observing the person and appropriately outputs very high uncertainties in these cases, which we show below. However, including these time points would swamp the influence of adding cameras.

As expected, our method showed tighter confidence intervals on the joint angles with more cameras. This was a fairly small effect when analyzing the median errors. However, with fewer cameras, there are also more periods with body parts occluded or out of view, and thus, the uncertainty about particular joints at certain times will be much greater than this median value, particularly when not analyzing only the carefully selected time points as we used in this analysis.

\begin{figure}[!htbp]
\centering
\includegraphics[width=1\linewidth]{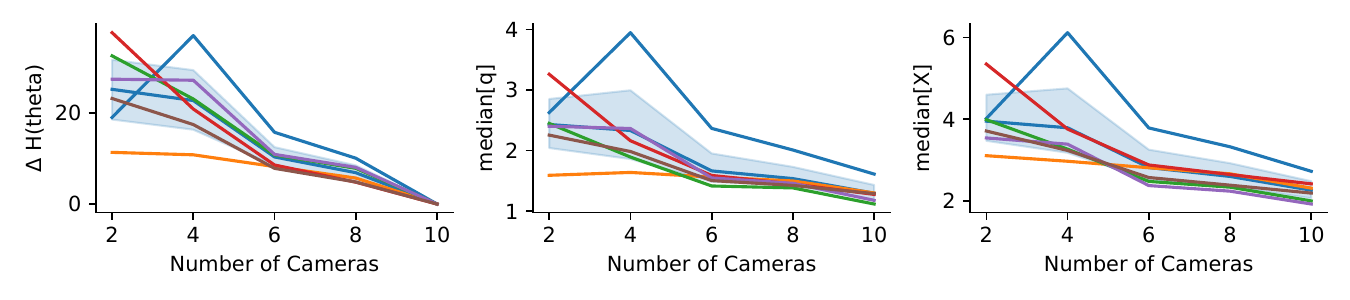}
\caption[]{The influence of removing cameras. The left panel shows the distribution entropy relative to the full camera complement. The middle shows the median value for the joint angle uncertainty. The right panel shows the median spatial error.}
\label{fig:remove_cameras}
\end{figure}

\subsection{Population Metrics}

After performing these optimizations, we applied this method to 2957 trials in our database from 264 individuals over 381 sessions. This dataset was intentionally heterogeneous. It included varying the number of cameras and their geometry. The population also included some able-bodied individuals (41) and people with gait impairments from various etiologies, including prosthesis users (74) and individuals with neurologic conditions. It included a variety of mobility assessments in addition to walking, such as the timed-up-and-go test and four-square-step test. The setting varied from data collection at conferences to outpatient and inpatient clinics, including a system integrated into the therapy gym used during inpatient rehabilitation. In many cases, multiple additional people, such as physical therapists or assistants, were the subject of interest.  As designed, this method identified some trials where tracking quality was poor due to occlusions or a limited number of views. It also showed that, in general, our tracking accuracy is very good.

\subsubsection{Spatial Errors}

\begin{figure}[!htbp]
\centering
\includegraphics[width=1\linewidth]{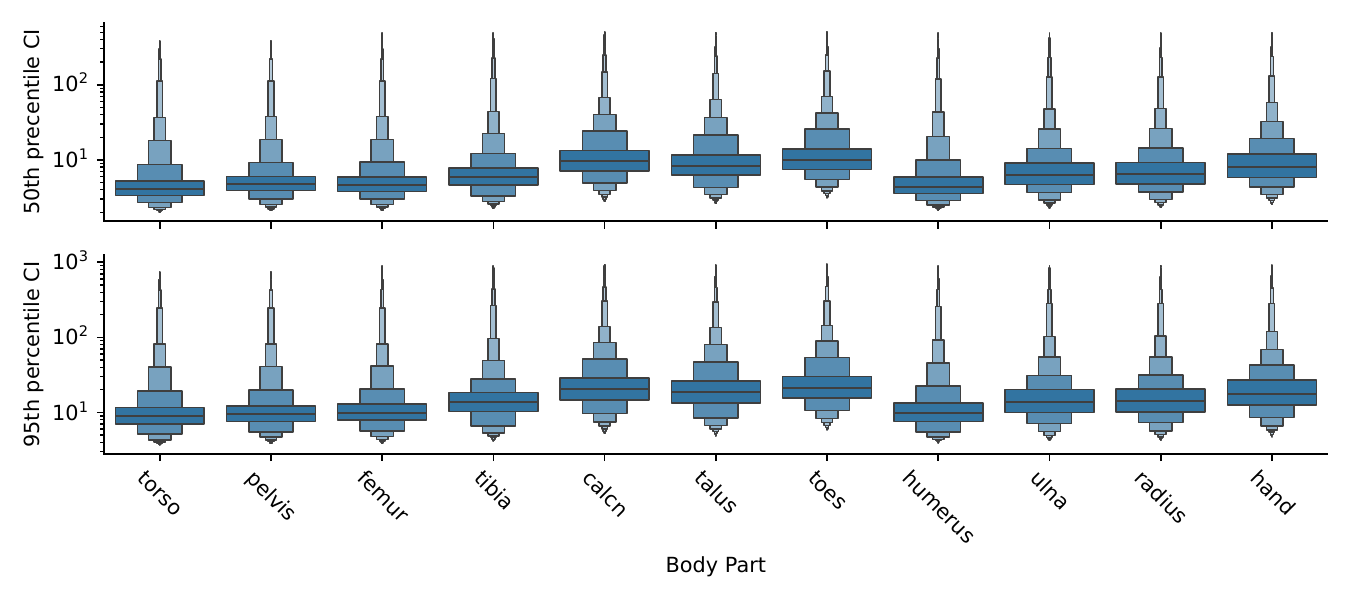}
\caption[]{The top row shows the 50th percentile spatial confidence interval for the body parts. The second row shows the 95th percentile confidence interval.}
\label{fig:body_position_noise}
\end{figure}

We quantified the spatial error by drawing 250 samples from our probability distribution at each time point and passing them through the forward kinematic model. For each sample, we then computed the Euclidean distance between each body segment position from the geometric median. At each time point, we computed the 50th (median) and 95th percentile error. We then took the average over time for those body segments. These body segments correspond to the bony elements of our model and not the virtual marker locations. The histogram of these errors for each joint is shown in Figure~\ref{fig:body_position_noise}. The 95th percentile of these errors across all joints and subjects was about 20mm, which corresponds closely with the error we saw in our prior validation studies against an instrumented walkway for spatial gait parameters \citep{cotton_optimizing_2023, cotton_differentiable_2024}.

For the sake of the visualization scale, we excluded 131 of our 2957 trials that had an average 95th percentile across the joints exceeding 1 meter. These typically occur in situations of significant occlusions or problems with the initial camera calibration. 

\subsubsection{Joint Angle Errors}

\begin{figure}[!htbp]
\centering
\includegraphics[width=1\linewidth]{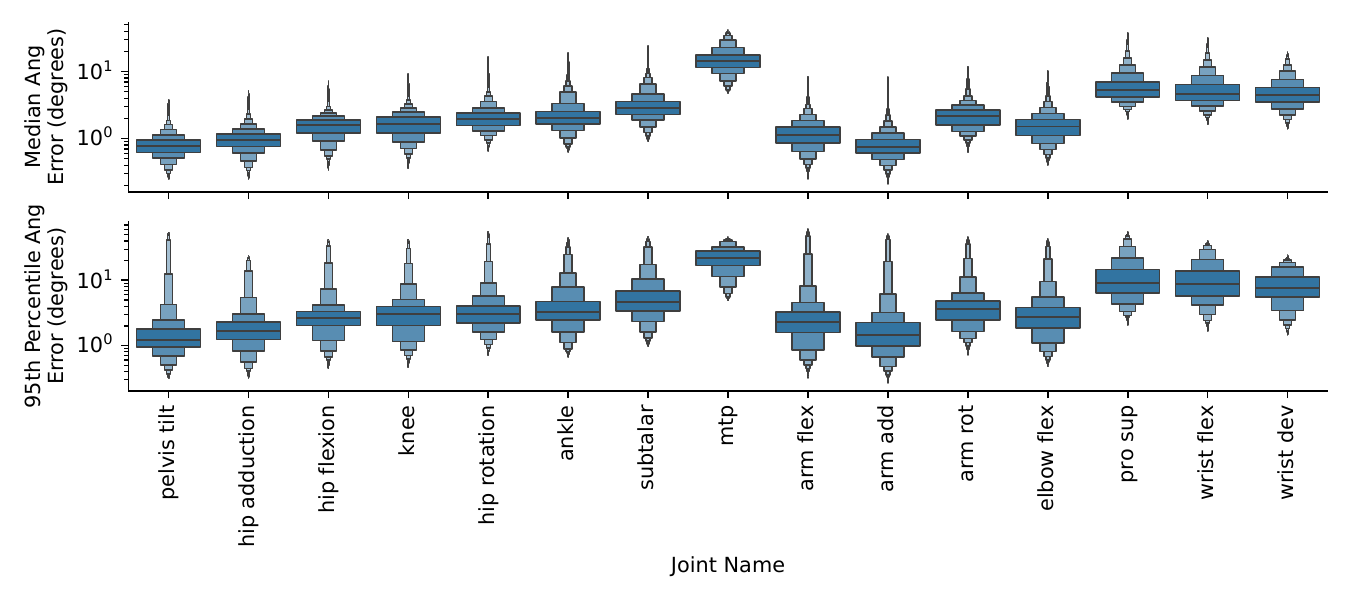}
\caption[]{The top row shows the 50th percentile joint angle error interval for the joints. The second row shows the 95th percentile error over the time series.}
\label{fig:joint_angle_noise}
\end{figure}

To quantify the joint angle errors, we took the standard deviation for each joint at each point of time from our variational posterior. We then took the median and 95th percentile over time for a given trial. The histogram of these broken down by joint is shown in Figure~\ref{fig:joint_angle_noise}. We saw greater errors in more distal joints, such as the ankle. We also saw that the parameters of the distal arm are poorly tracked. We note this was not using detailed hand keypoints we used in other work, which can reconstruct the hand within several mm \citep{firouzabadi_biomechanical_2024}. The median joint angle errors over time for the proximal joints were only a few degrees.

\subsubsection{Expected Calibration Error (ECE)}

We also computed population statistics on the ECE with the likelihood width clipped at 0, 1, and 2 pixels. Their 50th and 95th percentile values were respectively (0.33, 0.43), (0.07, 0.28), and (5e-4, 0.095). This suggests our vartional posterior width is sufficient to ``cover'' the majority of the keypoints with only 1 or 2 pixels of detection noise and further strongly suggests our distributions our fairly well calibrated and not overly confident about erroneous fits.

\subsubsection{Correlation Structure}

\begin{figure}[!htbp]
\centering
\includegraphics[width=1\linewidth]{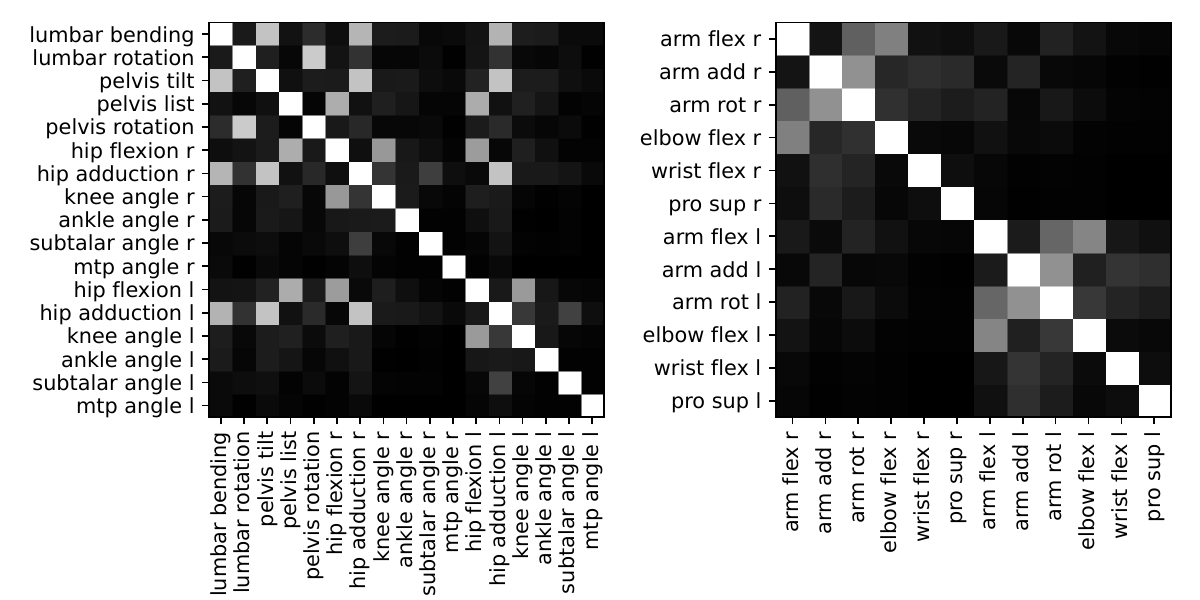}
\caption[]{The left panel shows the average correlation structure focused on joints in the lower body. The right panel shows the correlation structure for the upper body.}
\label{fig:joint_correlation_structure}
\end{figure}

Our method also captures the correlation structure of our pose uncertainty. We visualized this by averaging over time and then taking the median of the absolute value over trials. This is shown in Figure~\ref{fig:joint_correlation_structure}. Reassuringly, it captures numerous intuitively expected correlations between parameters such as lumber bending and hip flexion or between shoulder flexion and elbow flexion.

\subsubsection{Examples}

Here are two examples of the results of our method applied to walking trials. The top panel shows an example of a trial consistently tracked well throughout the room for an individual with an asymmetrical gait impairment from a stroke, zooming in to several gait cycles. Consistent with our statistics above, the ankle and subtalar joints are tracked more poorly than the hip and knee, although the asymmetry is visible. Furthermore, it is clear the uncertainty does still change differently for individual joints across the room. In the bottom panel, the tracking is good for most of the trial. Still, at the edge of the room, occlusions result in very large uncertainties over joint angles and position. This shows how we can track the accuracy of individual joints even through the room.

\begin{figure}[!htbp]
\centering
\includegraphics[width=1\linewidth]{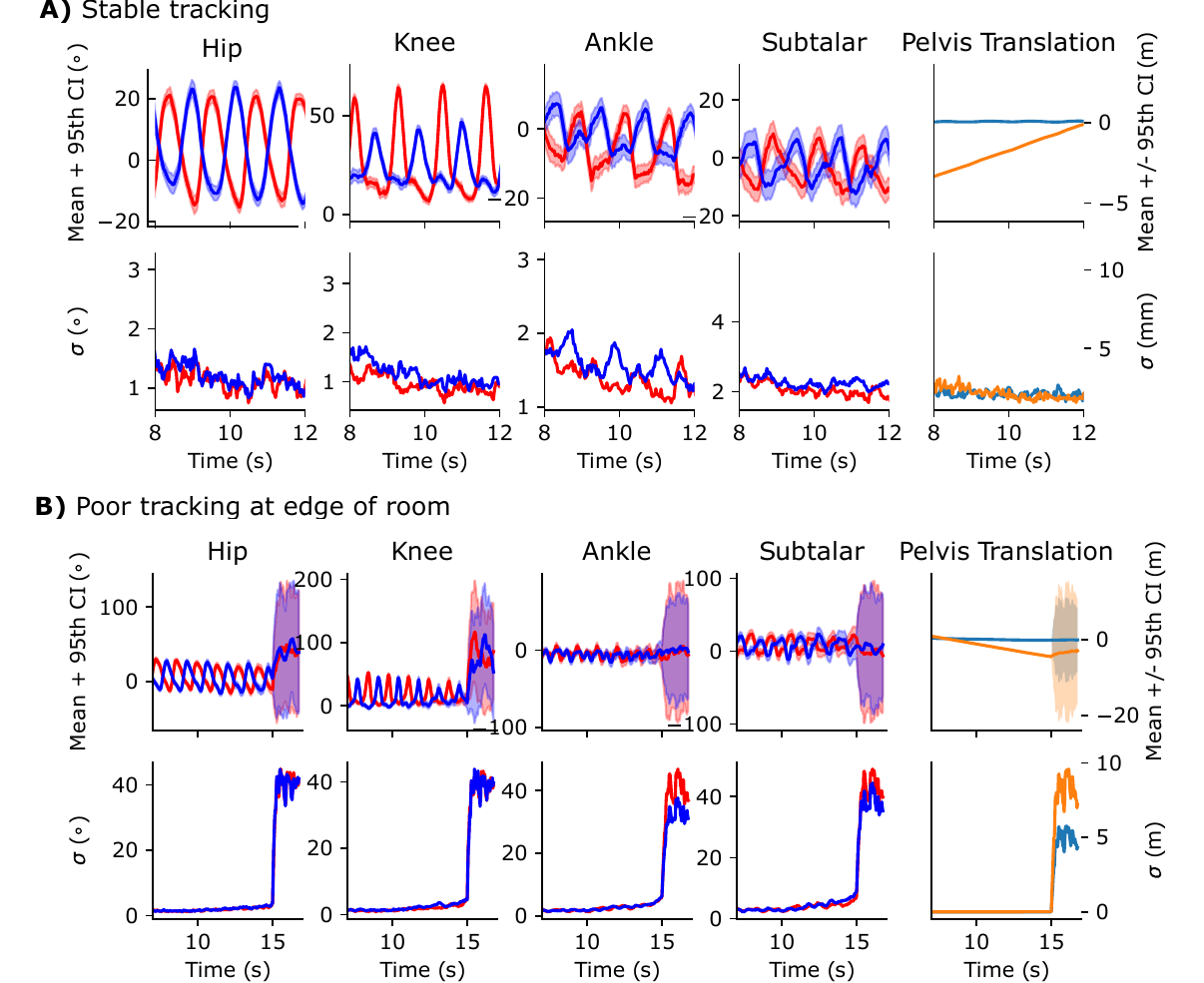}
\caption[]{Each subpanel shows the joint angle and position estimates with their 95th percentile confidence intervals (1.96 times the standard deviation from the Gaussian), and the second row in each panel shows standard deviations from the variational posterior. Note that in the bottom panel, the spatial error is shown in meters compared to the top panel, where it is shown in mm.}
\label{fig:tracking}
\end{figure}

\section{Discussion}

We showed how to compute a posterior probability over a kinematic pose trajectory from multiview markerless motion capture. This uses an implicit representation of the pose to represent the mean trajectory and covariance structure and optimizes this trajectory using variational inference. The optimization also jointly learns the likelihood function over keypoints, particularly how the likelihood function width scales with the keypoint score. At its core, this method leverages the constraints of biomechanics and multiview geometry to provide a signal for learning the confidence intervals consistent with the measurement discrepancies.

These confidence intervals are much more useful than the quality metrics we previously used, such as the mean reprojection error \citep{cotton_optimizing_2023, cotton_improved_2023, cotton_differentiable_2024}. They directly reflect what we want - how accurately we are tracking joint angles and body segment positions at each moment of time. In contrast to a traditional validation study aggregating statistics comparing two modalities in a particular condition, such as a camera geometry and clinical population, this method allows estimating the accuracy for a novel individual and camera configuration. Our fairly low ECE scores suggest our confidence intervals are fairly well calibrated, meaning 95\% of the time, the ground truth should fall within our 95\% confidence intervals. This is further supported by the fact that our median spatial error for the calcaneus (typically less than 10 mm) is similar to our stride length error compared to an instrumented walkway (also typically less than 10 mm) \citep{cotton_differentiable_2024}.  Our lesion experiments provide additional evidence for the power of this method, such as recovering the amount of injected keypoint noise and increasing the uncertainty with reducing the number of cameras.

We are also optimistic that this method will be a powerful tool for evaluating the accuracy of MMMC pipelines, including testing newer keypoint detectors and determining how accurate their reliability metrics are. We also found some indications that it is less inclined to fall into local minima than our prior deterministic method, as indicated by fewer fits with very high reprojection errors. Inconsistencies between the views occurring at these local minima should result in greater uncertainty, allowing variational sampling to identify more plausible fits. However, our method also has limitations. Optimizing the probabilistic representation is several times slower than our non-probabilistic implicit representation; for example, 10-20 trials can take 1-2 hours on an A100. Our synthetic testing also revealed that certain modeling discrepancies, such as errors in the calibration, can produce overly confident results by attributing too much uncertainty to the likelihood function. Our ECE metrics also do not show perfectly calibrated posterior distributions. Early experiments suggest the ECE can be incorporated into the loss to further prevent overconfidence, and we defer exploring this more comprehensively to future work.

Our results, particularly the calibration accuracy, would benefit from external validation using a marker-based method, although these produce near-ground-truth marker locations rather than poses, which depend on modeling assumptions.  There may be systematic biases between markerless and marker-based methodologies, with the latter not being a perfect ground truth representation of the bone movements due to factors such as marker errors and soft tissue movement artifacts. Our method also determines the uncertainty consistent with the modeling assumptions and the biomechanical model used. Thus, factors like viewpoint-dependent errors in the keypoint locations and confidence with occlusions and limited views may produce some biased results while being overconfident. We expect additional detailed analysis of residual errors to identify these problems. Ultimately, we hope for datasets containing synchronized video and bi-planar fluoroscopy that would provide a source of ground truth.

\section{Conclusion}

Reconstructing biomechanical trajectories with confidence intervals from multiview markerless motion capture can be accomplished using variational inference, where inconsistencies across views or time provide a learning signal for when confidence intervals should be wider. This method goes beyond traditional validation studies that are only designed to answer which metrics perform well on average under the validated condition. Instead, it estimates the joint angle and position confidence intervals for a unique individual and camera geometry, possibly in the presence of multiple occlusions. Our confidence intervals provide a critical signal for when MMMC results can be trusted for further analysis, which is essential for scaling to clinical practice or big-data movement science.

\section*{Acknowledgments}
RJC is supported by the Craig H. Neilsen Foundation, the Restore Center P2C (NIH P2CHD101913), R01HD114776, and the Research Accelerator Program of the Shirley Ryan AbilityLab. FHS was supported by the German Research Foundation (DFG): SFB 1233, Robust Vision: Inference Principles and Neural Mechanisms Project-ID 276693517, and SFB 1456, Mathematics of Experiment -- Project-ID 432680300. Thanks to the Cotton Lab team for help with data collection and discussion of the algorithms. Thanks to the individuals who have participated in our data collections and the therapists that were helping some of those patients. 

\bibliographystyle{unsrtnat}
\bibliography{main.bib}

\begin{thebibliography}{19}
\providecommand{\natexlab}[1]{#1}
\providecommand{\url}[1]{\texttt{#1}}
\expandafter\ifx\csname urlstyle\endcsname\relax
  \providecommand{\doi}[1]{doi: #1}\else
  \providecommand{\doi}{doi: \begingroup \urlstyle{rm}\Url}\fi

\bibitem[Cotton(2024)]{cotton_differentiable_2024}
R.~James Cotton.
\newblock Differentiable {Biomechanics} {Unlocks} {Opportunities} for {Markerless} {Motion} {Capture}, 2 2024.
\newblock [Online; accessed 2024-02-28].

\bibitem[Todorov et~al.(2012)Todorov, Erez, and Tassa]{todorov_mujoco_2012}
Emanuel Todorov, Tom Erez, and Yuval Tassa.
\newblock Mujoco: A {Physics} {Engine} for {Model}-{Based} {Control}.
\newblock In \emph{2012 {IEEE}/{RSJ} {International} {Conference} on {Intelligent} {Robots} and {Systems}}, pages 5026--5033, 10 2012.
\newblock \doi{10.1109/IROS.2012.6386109}.
\newblock [Online; accessed 2023-11-06].

\bibitem[Caggiano et~al.(2022)Caggiano, Wang, Durandau, Sartori, and Kumar]{caggiano_myosuite_2022}
Vittorio Caggiano, Huawei Wang, Guillaume Durandau, Massimo Sartori, and Vikash Kumar.
\newblock Myosuite -- {A} {Contact}-{Rich} {Simulation} {Suite} for {Musculoskeletal} {Motor} {Control}, 5 2022.
\newblock [Online; accessed 2023-12-28].

\bibitem[Hartley and Zisserman(2003)]{hartley_multiple_2003}
Richard Hartley and Andrew Zisserman.
\newblock \emph{Multiple {View} {Geometry} in {Computer} {Vision}}.
\newblock Cambridge University Press, 2003.
\newblock ISBN 978-0-521-54051-3.

\bibitem[Kidger and Garcia(2021)]{kidger_equinox_2021}
Patrick Kidger and Cristian Garcia.
\newblock Equinox: Neural {Networks} in {JAX} via {Callable} {PyTrees} and {Filtered} {Transformations}.
\newblock \emph{Differentiable Programming workshop at Neural Information Processing Systems 2021}, 10 2021.
\newblock [Online; accessed 2021-11-02].

\bibitem[Kingma and Welling(2014)]{kingma_auto-encoding_2014}
Diederik~P. Kingma and Max Welling.
\newblock Auto-{Encoding} {Variational} {Bayes}.
\newblock In \emph{2nd {International} {Conference} on {Learning} {Representations}, {ICLR} 2014 - {Conference} {Track} {Proceedings}}. International Conference on Learning Representations, ICLR, dec 20 2014.

\bibitem[Loshchilov and Hutter(2019)]{loshchilov_decoupled_2019}
Ilya Loshchilov and Frank Hutter.
\newblock Decoupled {Weight} {Decay} {Regularization}, 1 2019.
\newblock [Online; accessed 2023-07-02].

\bibitem[{DeepMind} et~al.(2020){DeepMind}, Babuschkin, Baumli, Bell, Bhupatiraju, Bruce, Buchlovsky, Budden, Cai, Clark, Danihelka, Dedieu, Fantacci, Godwin, Jones, Hemsley, Hennigan, Hessel, Hou, Kapturowski, Keck, Kemaev, King, Kunesch, Martens, Merzic, Mikulik, Norman, Papamakarios, Quan, Ring, Ruiz, Sanchez, Sartran, Schneider, Sezener, Spencer, Srinivasan, Stanojevi{\' c}, Stokowiec, Wang, Zhou, and Viola]{deepmind2020jax}
{DeepMind}, Igor Babuschkin, Kate Baumli, Alison Bell, Surya Bhupatiraju, Jake Bruce, Peter Buchlovsky, David Budden, Trevor Cai, Aidan Clark, Ivo Danihelka, Antoine Dedieu, Claudio Fantacci, Jonathan Godwin, Chris Jones, Ross Hemsley, Tom Hennigan, Matteo Hessel, Shaobo Hou, Steven Kapturowski, Thomas Keck, Iurii Kemaev, Michael King, Markus Kunesch, Lena Martens, Hamza Merzic, Vladimir Mikulik, Tamara Norman, George Papamakarios, John Quan, Roman Ring, Francisco Ruiz, Alvaro Sanchez, Laurent Sartran, Rosalia Schneider, Eren Sezener, Stephen Spencer, Srivatsan Srinivasan, Milo{\v s} Stanojevi{\' c}, Wojciech Stokowiec, Luyu Wang, Guangyao Zhou, and Fabio Viola.
\newblock The {DeepMind} {JAX} {Ecosystem}, 2020.

\bibitem[Dillon et~al.(2017)Dillon, Langmore, Tran, Brevdo, Vasudevan, Moore, Patton, Alemi, Hoffman, and Saurous]{dillon_tensorflow_2017}
Joshua~V. Dillon, Ian Langmore, Dustin Tran, Eugene Brevdo, Srinivas Vasudevan, Dave Moore, Brian Patton, Alex Alemi, Matt Hoffman, and Rif~A. Saurous.
\newblock Tensorflow {Distributions}, 11 2017.
\newblock [Online; accessed 2024-03-05].

\bibitem[Werling et~al.(2022)Werling, Raitor, Stingel, Hicks, Collins, Delp, and Liu]{werling_rapid_2022}
Keenon Werling, Michael Raitor, Jon Stingel, Jennifer~L. Hicks, Steve Collins, Scott~L. Delp, and C.~Karen Liu.
\newblock Rapid {Bilevel} {Optimization} to {Concurrently} {Solve} {Musculoskeletal} {Scaling}, {Marker} {Registration}, and {Inverse} {Kinematic} {Problems} for {Human} {Motion} {Reconstruction}, 8 2022.
\newblock [Online; accessed 2022-09-29].

\bibitem[Song et~al.(2019)Song, Diethe, Kull, and Flach]{song_distribution_2019}
Hao Song, Tom Diethe, Meelis Kull, and Peter Flach.
\newblock Distribution {Calibration} for {Regression}.
\newblock In \emph{Proceedings of the 36th {International} {Conference} on {Machine} {Learning}}, pages 5897--5906. PMLR, 5 2019.
\newblock [Online; accessed 2024-03-03].

\bibitem[Pierzchlewicz et~al.(2023)Pierzchlewicz, Bashiri, Cotton, and Sinz]{pierzchlewicz_optimizing_2023}
Pawe\l{}~A. Pierzchlewicz, Mohammad Bashiri, R.~James Cotton, and Fabian~H. Sinz.
\newblock Optimizing {MPJPE} {Promotes} {Miscalibration} in {Multi}-{Hypothesis} {Human} {Pose} {Lifting}.
\newblock In \emph{ICLR 2023 {Affinity} {Workshop}}, 4 2023.
\newblock [Online; accessed 2023-05-30].

\bibitem[Dong et~al.(2021)Dong, Fang, Jiang, Yang, Bao, and Zhou]{dong2021fast}
Junting Dong, Qi~Fang, Wen Jiang, Yurou Yang, Hujun Bao, and Xiaowei Zhou.
\newblock Fast and {Robust} {Multi}-{Person} 3d {Pose} {Estimation} and {Tracking} from {Multiple} {Views}.
\newblock In \emph{T-{PAMI}}, 2021.

\bibitem[S{\' a}r{\' a}ndi et~al.(2023)S{\' a}r{\' a}ndi, Hermans, and Leibe]{sarandi_learning_2023}
Istv{\' a}n S{\' a}r{\' a}ndi, Alexander Hermans, and Bastian Leibe.
\newblock Learning 3d {Human} {Pose} {Estimation} from {Dozens} of {Datasets} {Using} a {Geometry}-{Aware} {Autoencoder} to {Bridge} {Between} {Skeleton} {Formats}.
\newblock In \emph{IEEE/{CVF} {Winter} {Conference} on {Applications} of {Computer} {Vision} ({WACV})}. arXiv, 2023.
\newblock \doi{10.48550/arXiv.2212.14474}.
\newblock [Online; accessed 2023-01-02].

\bibitem[Ghorbani et~al.(2021)Ghorbani, Mahdaviani, Thaler, Kording, Cook, Blohm, and Troje]{ghorbani_movi_2021}
Saeed Ghorbani, Kimia Mahdaviani, Anne Thaler, Konrad Kording, Douglas~James Cook, Gunnar Blohm, and Nikolaus~F. Troje.
\newblock Movi: A {Large} {Multi}-{Purpose} {Human} {Motion} and {Video} {Dataset}.
\newblock \emph{PLOS ONE}, 16\penalty0 (6):\penalty0 e0253157, 6 2021.
\newblock ISSN 1932-6203.
\newblock \doi{10.1371/journal.pone.0253157}.
\newblock [Online; accessed 2023-02-20].

\bibitem[Cotton et~al.(2023{\natexlab{a}})Cotton, DeLillo, Cimorelli, Shah, Peiffer, Anarwala, Abdou, and Karakostas]{cotton_optimizing_2023}
R.~James Cotton, Allison DeLillo, Anthony Cimorelli, Kunal Shah, J.~D. Peiffer, Shawana Anarwala, Kayan Abdou, and Tasos Karakostas.
\newblock Optimizing {Trajectories} and {Inverse} {Kinematics} for {Biomechanical} {Analysis} of {Markerless} {Motion} {Capture} {Data}.
\newblock In \emph{IEEE {International} {Consortium} for {Rehabilitation} {Robotics}}. arXiv, 3 2023{\natexlab{a}}.
\newblock \doi{10.48550/arXiv.2303.10654}.
\newblock [Online; accessed 2023-03-21].

\bibitem[Cotton et~al.(2023{\natexlab{b}})Cotton, Cimorelli, Shah, Anarwala, Uhlrich, and Karakostas]{cotton_improved_2023}
R.~James Cotton, Anthony Cimorelli, Kunal Shah, Shawana Anarwala, Scott Uhlrich, and Tasos Karakostas.
\newblock Improved {Trajectory} {Reconstruction} for {Markerless} {Pose} {Estimation}.
\newblock In \emph{45th {Annual} {International} {Conference} of the {IEEE} {Engineering} in {Medicine} and {Biology} {Society}}, Sydney, 3 2023{\natexlab{b}}.
\newblock [Online; accessed 2023-03-09].

\bibitem[Biewald(2020)]{wandb}
Lukas Biewald.
\newblock Experiment {Tracking} with {Weights} and {Biases}, 2020.

\bibitem[Firouzabadi et~al.(2024)Firouzabadi, Murray, Sobinov, Peiffer, Shah, Miller, and Cotton]{firouzabadi_biomechanical_2024}
Pouyan Firouzabadi, Wendy Murray, Anton~R Sobinov, J.D. Peiffer, Kunal Shah, Lee~E Miller, and R.~James Cotton.
\newblock Biomechanical {Arm} and {Hand} {Tracking} with {Multiview} {Markerless} {Motion} {Capture}.
\newblock In \emph{2024 10th {IEEE} {RAS}/{EMBS} {International} {Conference} for {Biomedical} {Robotics} and {Biomechatronics} ({BioRob})}, pages 1641--1648, 9 2024.
\newblock \doi{10.1109/BioRob60516.2024.10719940}.
\newblock URL \url{https://ieeexplore.ieee.org/document/10719940}.
\newblock [Online; accessed 2024-10-30].

\end{thebibliography}

\section*{Appendix}

\subsection*{Metric: Expected Calibration Error}

A key motivation of this work is to provide confidence intervals for our pose estimates to indicate when our data can be trusted or not. These confidence intervals must be well-calibrated. For example, if we estimate a 95\% confidence interval, the ground truth should fall within this interval approximately 95\% of the time. This can be quantified with the Expected Calibration Error (ECE) \citep{song_distribution_2019, pierzchlewicz_optimizing_2023}, which measures the agreement between predicted confidence intervals and the actual empirical distribution of errors. However, we do not have the true pose to measure or validate this. Obtaining ground truth would require using biplanar fluoroscopy to track bones in real-time, which is impractical at scale.

Instead, we assess calibration by comparing our predicted keypoint distributions to the detected keypoints in image space. The keypoint distribution has two sources of uncertainty. First, the epistemic uncertainty in the pose trajectory is modeled by our variational posterior $q(\boldsymbol \theta_t)$, which maps to a distribution over predicted keypoints. The second source of uncertainty is the measurement uncertainty from the keypoint detector, which is captured by the likelihood function. The combination of these sources should reflect the true variability in keypoint detections. However, a potential failure mode arises if the variational posterior underestimates uncertainty while the likelihood function overestimates it, creating calibrated keypoint confidence intervals that do not accurately represent pose uncertainty.

We mitigate this failure mode when computing the calibration of the keypoints through two tweaks. First, we approximate ground-truth keypoints using the 5th percentile of keypoints with the highest detection confidence. These should have the least measurement noise, allowing them to serve as pseudo-ground truth for evaluating calibration. Furthermore, we clip the width of the likelihood function, $\sigma=\min(\sigma_\psi(s), \sigma_{\mathrm ECE})$, independently of the values of $\psi$ and $s$, preventing excess uncertainty from being attributed to the likelihood function. Intuitively, for the limit of true ground truth keypoints, using a $\sigma_{\mathrm ECE}=0$, a calibrated result would require that the uncertainty for the keypoints induced by the posterior $q^{\hat{\mathbf Y_c}}(\hat{\mathbf y_c}) = (\Pi_c \mathcal M_\beta)_{*}q(\boldsymbol \theta)$ is sufficiently broad to cover all the keypoints from all the views, where $(\Pi_c \mathcal M_\beta)_{*}$ is the pushforward operator that maps from pose densities to keypoint densities. In practice, we monitor the ECE for several $\sigma_{\mathrm ECE}$ values from 0 to 5 pixels. We implement this by using a Jacobian of the pushforward operator to build a Taylor approximation of $q^{\hat{\mathbf Y_c}}(\hat{\mathbf y_c})$ and then the final distribution has a variance of $\sigma_{\mathrm ECE}^2$ added to it. This produces a 2D distribution for the keypoint location with a mean location, $\mu$, and scale, $\sigma^2$. While strictly this can be anisotropic, we use the average scale over the x and y directions. Thus, for each of the keypoints we are using for the ECE, we have a predicted distribution, which we denote as $D_i$, and a detected location $\mathbf y_i$.

Instead of binning the predicted error quantiles, we compute the ECE through an equivalent formulation using a goodness of fit test called a probability-probability (p-p) plot, where the empirical cumulative distribution of keypoint errors is plotted against the predicted cumulative distribution. This can also be understood as an application of the Probability Integral Transform (PIT), where a random variable $X$ is transformed through its cumulative distribution function F, producing a transformed variable $C = F(X)$ that is uniformly distributed on the interval [0,1].  The rank-ordered samples of $C$ should then fall on the diagonal line. To apply this to samples of $(\mathbf y_i, D_i)$ we first measure the radial error $\text{err}_i = \|\mathbf y_i - \mu_i\|$. Radial distances for isotropic 2D Gaussians are expected to follow a Rayleigh distribution, which we use in our CDF: $c_i = F_{\mathrm {Rayleigh}(\sigma)}(\text{err}_i)$. We then sort the sample samples of $\{c_i\}$ in ascending order to obtain $\tilde{c}_1 \le \dots \le \tilde{c}_N$. Finally, we compute the
$\mathrm{ECE} \;=\; \frac{1}{N}\sum_{i=1}^N \bigl|\tilde{c}_i - p_i\bigr|$, where  $\{p_i = i/N\}$. When explicitly noting the $\sigma_{\mathrm ECE}$ used to compute $D_i$, we note the ECE as $\mathrm{ECE}_{\sigma_{\mathrm ECE}}$.

\subsection*{Taylor approximation of $q^{\hat{\mathbf Y}_c}(\hat{\mathbf y}_c)$}

While $q^{\hat{\mathbf Y_c}}(\hat{\mathbf y_c})$ cannot be analytically computed from $q(\boldsymbol \theta)$, we can approximate it using a Taylor expansion of the transformation from pose space to the image plane. $\mathcal M_\beta$ maps from pose space to site markers in 3D space, which we can then project into the image plane using the calibrated camera model, $\Pi_c:  \mathbf x_t \in \mathbb R^{J \times 3} \rightarrow \mathbf y_t^c \in \mathbb R^{J \times 2}$, with $c$ indexing over cameras. This gives us a transformation from pose space to image space:
\begin{equation*}
\label{eq:pose_to_image}
\hat {\mathbf y}_t^c = g_c(\boldsymbol \theta_t) = \Pi_c \, \mathcal M_\beta \, \boldsymbol \theta_t
\end{equation*}
The first-order Taylor expansion of $g_c$ about the mean, $\boldsymbol \mu_{\boldsymbol \theta_t}=\boldsymbol \mu_{\phi}(t)$, is:
\begin{equation*}
g_c(\boldsymbol \theta_t) \approx g_c(\mu_{\boldsymbol \theta _t}) + J_g(\boldsymbol \mu_{\boldsymbol \theta_t})\left( \boldsymbol \theta_t - \boldsymbol \mu_{\boldsymbol \theta_t}\right)
\end{equation*}

With $J_g(\boldsymbol \theta_0)=\left.\frac{\partial g_c}{\partial \boldsymbol \theta}\right\vert_{\boldsymbol \theta=\boldsymbol \theta_0}$.
From this, the first and second moments of the predicted keypoint distribution after the pushforward operator, $q^{\hat{\mathbf Y_c}}(\hat{\mathbf y_c}) = (\Pi_c \mathcal M_\beta)_{*}q(\boldsymbol \theta)$:
\begin{equation*}
\label{eq:taylor_mean}
\mathbb E[\boldsymbol y_t^c]=g_c(\boldsymbol \mu _{\boldsymbol \theta_t})
\end{equation*}

\begin{equation*}
\label{eq:taylor_cov}
\mathrm{Cov}[\boldsymbol y_t^c] \approx \mathrm{Cov}[g(\boldsymbol \theta_t)]_{\boldsymbol \theta_t \sim q(\boldsymbol \theta_t)} \approx J_g(\boldsymbol \mu_{\boldsymbol \theta_t})\, \Sigma_{\boldsymbol \theta_t} \, J_g(\boldsymbol \mu_{\boldsymbol \theta_t})^\top
\end{equation*}

\end{document}